\documentclass[article]{elsarticle}

\usepackage{stmaryrd}
\usepackage{lineno,hyperref}
\usepackage[small]{caption}
\usepackage{amsmath}

\usepackage[font=small]{caption}

\usepackage{graphicx}
\usepackage{subfig}
\usepackage{color}

\usepackage{siunitx}

\DeclareMathOperator*{\argmax}{arg\,max}

\modulolinenumbers[5]

\journal{International Journal of Approximate Reasoning}

\begin{document}

\begin{frontmatter}

\title{Quick and energy-efficient Bayesian computing of binocular disparity using stochastic digital signals}

\author[isiraddress]{Alexandre Coninx\corref{mycorrespondingauthor}}
\cortext[mycorrespondingauthor]{Corresponding author}
\ead{alexandre.coninx@isir.upmc.fr}
\author[isiraddress]{Pierre Bessi\`{e}re}
\author[isiraddress]{Jacques Droulez}

\address[isiraddress]{ISIR CNRS/UPMC, 4 place Jussieu 75005 Paris, France}

\begin{abstract}
Reconstruction of the tridimensional geometry of a visual scene using the binocular disparity information is an important issue in computer vision and mobile robotics, which can be formulated as a Bayesian inference problem. However, computation of the full disparity distribution with an advanced Bayesian model is usually an intractable problem, and proves computationally challenging even with a simple model. In this paper, we show how probabilistic hardware using distributed memory and alternate representation of data as stochastic bitstreams can solve that problem with high performance and energy efficiency. We put forward a way to express discrete probability distributions using stochastic data representations and perform Bayesian fusion using those representations, and show how that approach can be applied to diparity computation. We evaluate the system using a simulated stochastic implementation and discuss possible hardware implementations of such architectures and their potential for sensorimotor processing and robotics.
\end{abstract}


\end{frontmatter}


\section{Introduction}
\label{sec:intro}
Using two cameras in a stereoscopic setup to reconstruct the tridimensional geometry of a visual scene, in a way similar to that performed by human stereopsis, is an important issue in computer vision, with major applications to autonomous robotics (and more specifically autonomous driving \citep{Geiger2012}). That issue has been an active research topic since at least 40 years, and a wide range of methods and algorithms have been proposed \citep{Scharstein2002,Lazaros2008} and evaluated on standardized benchmarks \citep{Geiger2013,Scharstein2014}.

Several works have shown that the binocular disparity computation can efficiently be formulated as a Bayesian inference problem \citep{Belhumeur1996,Su2012}. The disparity value for each pixel is then expressed as a discrete probability distribution, which can be computed through a probabilistic model using likelihood values specified from the image data. However, computing the full disparity distribution on whole images proves challenging and compuationally demanding. That's why most works on binocular disparity using Bayesian models instead reduce the output to a single disparity value per pixel (often using the maximium a-posteriori likelihood estimator). That approach simplifies the computation and allows to reformulate it as an energy minimization problem that can be solved efficiently by classic optimization techniques such as dynamic programming \citep[see][for an exemple]{Belhumeur1996}.

However, it means that although the computation is based on a probabilistic formalism, it yields deterministic disparity values and not disparity distributions, despite the latter representation being richer and offering many benefits, especially for robotics and sensorimotor systems. Full disparity distributions can accurately represent cases where stereopsis is not sufficient to completely determinate the world geometry, such as ambiguous pixels with multiple matches, or pixels with no matches (e.g. due to occlusions). Such probabilistic representations can also directly be used by Bayesian mapping and navigation methods such as the Bayesian occupation filter \citep{Coue2006}, and more generally by probabilistic and Bayesian robotics techniques \citep{Thrun2005,Lebeltel2006,Bessiere2008}. Bayesian inference also provides a powerful framework to express assumptions and prior knowledge about the structure of the world (for example the location of the ground or other known objects) as prior probability distributions.

Stochastic computing is a field dedicated to designing and using computing devices that are intentionally stochastic to perform probabilistic reasoning, using non-Von Neumann architectures, distributed memory and specific data representations. More specifically, the \href{https://www.bambi-fet.eu/}{BAMBI project} is a research effort to develop stochastic machines implementing Bayesian inference (Bayesian machines) \citep{Alves2015}. In this paper, we show how those Bayesian machines can be used to efficiently compute full binocular disparity distribution, paving the way towards fully stochastic autonomous robots and other sensorimotor systems.

In the remainder of this article, we will first give an overview of the related work in section~\ref{sec:sota}, both about stochastic computing and fast binocular disparity computation. We will then describe our Bayesian binocular disparity computation model in section~\ref{sec:disparity}. Section~\ref{sec:stochastic} will be dedicated to the description of the stochastic computer implementing that model, focusing first on the general principles of computation using stochastic bitstream and second to their application to the Bayesian disparity computation. The evaluation of that system and its results will be presented in section~\ref{sec:eval} and further discussed in section~\ref{sec:discu}. We will then conclude in section~\ref{sec:ccl} by summing up the implications of that work for the design of Bayesian robotic systems using stochastic components and discussing the future prospects of that topic.

\section{Previous work}
\label{sec:sota}
\subsection{Hardware stochastic computing}

The general idea of stochastic computations with temporal coding can be traced back to the seminal works of \citet{VonNeumann1956} and \citet{Gaines1969} who highlighted the interest of such data representations, but their approaches were not widely pursued due to the rapid development of more efficient deterministic computers. The topic has recently received a renewed attention due to the development of probabilistic and Bayesian models in computer science and engineering -- and more specifically for sensorimotor and cognitive systems -- and the limitations of classic computers to implement those models.

The idea of developing hardware dedicated to bayesian reasoning has recently been pursued by several teams \citep{Vigoda2003,Mansinghka2009,Jonas2014}, exploring different computational paradigms to perform probabilistic inference. To address the problem of approximate inference \citet{Mansinghka2009} uses sampling methods for approximate inference and in a similar way, Jonas designed Markov Chain Monte Carlo based algorithms to provide a representation of probability distributions as sets of samplers \cite{Jonas2014}. To compute exact inference, a number of different frameworks and toolsets have been put forward. \citet{Vigoda2003} designed architectures based on probabilities represented by analog signals, and used the message passing algorithm to compute exact inference. More recently, a research project conducted at the Nanoscale Computing Fabrics Laboratory has led to the design of an unconventional hardware architecture based on electro-magnetic computations to perform inference on Bayesian Network models \citep{Khasanvis2015}. \citet{Ferreira2015} also showed that exact inference can be efficiently computed using GPU hardware for some high-dimensional problems. Finally, the approach taken by \citet{Thakur2016} is quite similar to ours: they use stochastic bitstreams and target special inference problems. They have proposed two frameworks, BEAST (Bayesian Estimation And Stochastic Tracker) and BIND (Bayesian INference in DAG), to perform inference using stochastic electronics on two types of Bayesian models, Hidden Markov Models and Direct Acyclic Graphs (DAG) respectively.

In the framework of the BAMBI project, another stochastic architecture has been proposed to perform naive Bayesian fusion using Muller C-Elements \citep{Friedman2016}, which achieves exact inference with normalization for binary random variables, but create harmful correlations in the stochastic signals and can't be easily extended to non-binary discrete distributions. Other recent work conducted within the BAMBI project have proposed using digital signals with temporal coding to perform Bayesian inference, and a proof-of-concept to solve a simple sensorimotor problem has been put forward \citep{Faix2015}. In this paper, we apply the same principles to a more computationally challenging Bayesian model to highlight their benefits.

\subsection{Disparity computation}

As it provides a way to estimate the depth information using data from standard digital cameras, the binocular disparity problem has received a wide attention since the beginnings of computer vision. Existing approaches have been summarized in reviews \citep{Scharstein2002,Lazaros2008}, which show that most methods follow the same general structure which can be divided in three steps:
\begin{enumerate}
\item Computing a \emph{matching cost}, which is a positive value associated to each possible pair of matching pixels\footnote{Most algorithms use rectified image pairs, which allows to only consider pixels on corresponding rows for matching, and limit the disparity to a maximum value $D_{max}$ corresponding to a minimum distance. $D_{max}$ depends on image resolution, camera focal length and visual environment; typical values are 50 to 100 pixels.}. The matching cost is a dissimilarity measure: the least likely the pixels are to match, the higher it is. The cost is computed locally, typically by comparing the luminance or color of individual pixels. The most common matching cost is the squared difference of pixel values \citep{Scharstein2002}, but some other techniques preprocess the image with operators such as the gradient \citep{Scharstein1994} or use banks of linear spatial filters \citep{Jones1992}.
\item Applying an optional \emph{cost aggregation}, which performs spatial integration of the pixel-wise information provided by cost values. The main goal of that step is to take into account the fact that most points of the disparity map are locally smooth and therefore neighbouring pixels have correlated disparity values. The simplest form of cost aggregation relies on averaging cost values for a given disparity across a given neighborhood.
\item An \emph{optimization} step, which uses the (aggregated) cost to compute the final disparity image. This step can be limited to simply selecting the disparity value associated to the lowest cost in a winner-takes-all way. But it can also involve global computations to optimize the disparity map with regard to a given world model (e.g. smoothness, plane surfaces, etc. \citep{Belhumeur1996}), using techniques such as dynamic programming, in which cases it can complement or replace cost aggregation.
\end{enumerate}

\subsubsection{Bayesian disparity computation}
\label{sssec:Bayesian-disp}
Several of the existing works \citep{Belhumeur1996,Su2012} use the Bayesian inference framework to describe this process. For example, \citet{Belhumeur1996} proposes to reconstruct the scene geometry $S$ from the left and right images $I_l$ and $I_r$ using a Bayesian model:

\begin{equation}
P(S|I_l, I_r) \propto P(S) \cdot P(I_l,I_r|S)
\label{eq:belhumeur}
\end{equation}
with $P(S)$ being a prior specifying the expected shape (smooth, etc.) of the world and $P(I_l,I_r|S)$ a data term computed from the matching cost. Computing $P(I_l,I_r|S)$ therefore corresponds to the matching cost computation step, there is no cost aggregation step, and computing and integrating the prior constitutes the optimization step. Belhumeur uses squared difference to compute the cost and proposes three increasingly complex world models to define the prior, but the computation of the full posterior probability distribution -- which has cardinality $(D_{max}+1)^{w \times h}$ -- is intractable.

He therefore uses an energy formalism and defines $E[S] = -\log(P(S) \cdot P(I_l,I_r|S))$, which allows to compute $\hat{S} = \argmax\limits_{S} P(S|I_l, I_r)$ by minimizing $E[S]$, and shows that a simplified form of this optimization problem can be solved by dynamic programming. As mentioned in section~\ref{sec:intro}, despite that algorithm being based on Bayesian inference, it only yields a single disparity value for each pixel.

\subsubsection{Supervised techniques}

The development of public image pairs datasets provided with a disparity baseline such as the KITTI dataset \citep{Geiger2013} or the Middlebury dataset \citep{Scharstein2014} have made it possible to treat disparity computation as a supervised machine learning problem. Some algorithms use deep convolutional networks to learn the matching cost \citep{Zbontar2014,Mayer2015}, and perform cost aggregation and optimization using other techniques.

Those techniques currently populate the top of the KITTI leaderboard\footnote{\url{http://www.cvlibs.net/datasets/kitti/eval_scene_flow.php?benchmark=stereo}, consulted 24/03/2016}. Although they are extremely accurate on benchmarks, their efficiency depend on the existence of a relevant supervised training dataset. Besides, they are computationally very intensive, using high-end CPUs and GPUs and sometimes requiring a computing time of several minutes per frame. Those features would make applying those techniques in a mobile robotics context challenging.

\subsubsection{Sampling approach}

An approach that is directly relevant to our positioning is the method proposed by \citet{Jonas2014} as an application of his aforementioned hardware architecture for approximate inference. In that work, they model the disparity distribution using a Markov random field, and use a hardware architecture using Gibbs sampling to sample the posterior distribution. Although this approach is efficient and allows to use a computationally intensive Bayesian disparity model with global optimization, it uses a unique, centralized pseudo-random number generator as source of entropy and lacks some of the features of our system, such as the high parallelism and the robust computation of the full disparity with a very low number of clock cycle.




\section{Bayesian disparity computation model}
\label{sec:disparity}
\subsection{Overview}
\begin{figure}
\centering
\includegraphics[width=.5\linewidth]{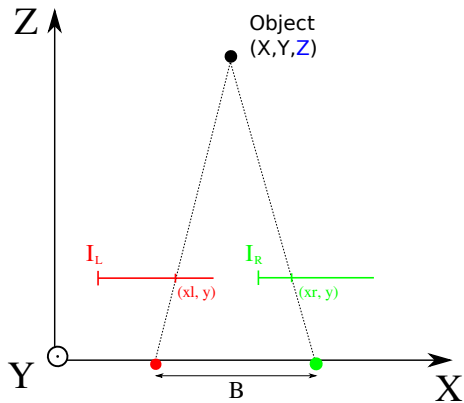}
\caption{Object observed by two cameras in a stereoscopic vision system. Matching corresponding pixels in the two images is key to computing the disparity $d$ and therefore the depth $Z$.}
\label{fig:disparity-schema}
\end{figure}

The goal of the disparity computation is to estimate the tridimensional geometry of a visual scene from two rectified images taken from two identical cameras with focal length $f$ distant from a known baseline distance $B$. If an object projects into the left camera's image plane $I_l$ at position $x$ and in the right camera's image plane $I_r$ at position $x - d$, its depth $Z$ from the cameras can be computed by $Z = \frac{B \cdot f}{d}$ (see fig.~\ref{fig:disparity-schema}). The goal of a disparity algorithm is therefore to identify matching pixels in the two images to compute the disparity.

\subsection{Model description}
\label{ssec:model}

As mentioned in section~\ref{sssec:Bayesian-disp}, the main obstacle to compute full disparity distributions is the very high cardinality of the considered distribution: integrating smoothness constraints in the probabilistic model requires to perform inference on distributions of size $(D_{max}+1)^{N_\textrm{pixels}}$, where $N_\textrm{pixels}$ is the number of pixels in the domain on which the optimization is performed. If the optimization is performed on the whole image or on entire rows or columns (as is the case in \citep{Belhumeur1996}) the problem becomes completely intractable, but even smaller integration neighborhoods are problematic. In order to avoid that issue, we will perform all of the spatial information integration as image preprocessing operations, and then only perform pixelwise Bayesian operations using the preprocessed data.

Our stereo matching method therefore relies on the preprocessing of images using linear convolution filters to extract relevant features. Other algorithms have used such convolution filters for disparity computation \citep{Jones1992}, although they process the feature information from those filter in a different way. The relevance of using such linear spatial filters as a preprocessing step is also highlighted by recent works using deep neural networks to compute disparity \citep{Zbontar2014,Mayer2015}, which use a convolutional layer (with filters trained through supervised learning) as their input.

The feature maps output by the filters are then used to compute feature matching costs for pixel pairs corresponding to the possible disparities. Those costs are used to compute probabilistic likelihood functions similar to those used by \citet{Belhumeur1996}, and those likelihood terms are then combined using naive Bayesian fusion.

In our method, we only use three simple square spatial filters of size 5 pixels to process images of width $W$ and height $H$:

\begin{itemize}
\item One simple luminance linear averaging filter $m$;
\item One linear horizontal luminance  gradient filter $g_H$;
\item One linear vertical luminance gradient filter $g_V$.
\end{itemize}

For each of the three filters $f \in \{m, g_H, g_V\}$, we compute the left and right feature maps by applying the filter to the left and right images: $f^l = I_l \ast f$ and $f^r = I_r \ast f$. Due to the size of the convolution filters, those feature maps have width $W_f = W-4$ and height $H_f = H-4$.

For each pair of feature maps and each possible disparity value we compute a matching cost, using the simple squared difference:

\begin{equation}
C_f(x,y,d) = (f^l(x,y) - f^r(x-d,y))^2
\label{eq:cost}
\end{equation}
for $d \in \llbracket 0 ; D_{max} \rrbracket$ and $x \geq D_{max}$.

The cost shown by Eq.~\ref{eq:cost} measures the dissimilarity between the pixels at coordinate $(x,y)$ in the left image and $(x-d,y)$ in the right image for the feature $f$. In order to use a Bayesian inference framework, we use these costs to compute likelihood probability values:

\begin{equation}
p(f^r(x-d,y)|f^l(x,y),[D(x,y)=d]) = p_0 + (1 - p_0)e^{-\frac{C_f(x,y,d)}{2\sigma_f^2}}
\label{eq:lh}
\end{equation}

Equation~\ref{eq:lh} expresses the likelihood of observing the value $f^r(x-d,y)$ in the right feature map if the value $f^l(x,y)$ is observed in the left feature map and the disparity at coordinates $(x,y)$ is $d$. That probabilistic formulation allows us to specify a base probability $p_0$ of the features matching even if the cost is high (which can happen when the two images locally differ for reasons unrelated to the problem, for example because of specular reflections), and a parameter $\sigma_f$ representing the expected inaccuracy of the cost measurement (a small value of $\sigma_f$ results in a null or very small cost being required to give a high likelihood value).

In the following, we will drop the $(x,y)$ and $(x-d,y)$ spatial coordinates in equations for better readability. Assuming conditional independance between the likelihood terms, we can compute the disparity distribution using naive Bayesian fusion:

\begin{equation}
p([D=d]|I_l,I_r) \propto p([D=d]) \prod\limits_{f \in \{m, g_H, g_V\}} p(f^r|f^l,[D=d])  
\label{eq:fusion}
\end{equation}
where $p(f^r|f^l,[D=d])$ are the the likelihoods computed by eq.~\ref{eq:lh} and $p(D)$ is a prior on the disparity distribution, which can either be set to uniform or be used to represent prior information about the world (for example, if we know the world contains a flat floor with no holes, the prior probability of disparities corresponding to objects under the floor can be set to zero). $p(D|I_l,I_r)$, is the posterior disparity distribution, which can be used in further probabilistic computation -- for example as input of a probabilistic occupancy model -- or estimated using the maximum a-posteriori (MAP) estimator: $d^* = \argmax\limits_{d \in \llbracket 0 ; D_{max} \rrbracket} p([D=d]|I_l,I_r)$

We described the algorithm with three simple filters operating on luminance data, but the same method can easily be extended to color processing (by applying each filter to each of the three color layers), or to using a higher number of filters using various convolution kernels.

\section{Stochastic implementation of the model}
\label{sec:stochastic}
Previous work \citep{Faix2015} has shown that naive Bayesian fusion could be performed by stochastic machines. In this section, we will describe that structure of a Bayesian machine architectured as a matrix of stochastic operators and explain how it can be used to implement the probabilistic binocular disparity computation detailed in section~\ref{ssec:model}.

\subsection{Stochastic Bayesian fusion}

\label{ssec:bm1}

\subsubsection{Probabilities as stochastic bitstreams}
\label{sssec:bitstreams}

Our stochastic computational architecture represents data using \textit{stochastic bitstreams}. Stochastic bitstreams are random digital binary signals that express a probability value (p-value)) by the proportion of bits set to 1 in a given signal (Fig.~\ref{sfig:bitstream}). Generating a stochastic bitstream $b$ encoding probability $p$ is therefore done by using a random number generator outputing random bits set to 1 with a probability $p$. Conversely, extracting the value of $p$ from $b$ and storing it as a floating point or fixed point number requires to integrate information from $b$ on an extended duration to count the proportion of bits set to 1, the precision of the recovered $p$ value increasing with the integration time.

If two probability values $p_1$ and $p_2$ are encoded by two uncorrelated stochastic bitstreams $b_1$ and $b_2$ and those two signals are input to a logic AND gate, the probability $p_{out}$ of the output signal $s_{out}$ to be in state 1 at a given time is given by :

\begin{align*}
p_{out} & = P([s_{out}=1]) \\
	& = P([s_1 = 1] \land [s_2 = 1]) \\
	& = P([s_1 = 1]) \cdot P([s_2 = 1]|[s_1 = 1]) \\
	& = P([s_1 = 1]) \cdot P([s_2 = 1]) \\
	& = p_1 \cdot p_2
\end{align*}
The stochastic signal data representation allows to perform probability product with a simple logic circuit.

\subsubsection{Representation of discrete random variables: the stochastic bus}
\label{sssec:bus}

A discrete random variable $V$ with cardinality $M$ can be represented by a set of $M$ stochastic bitstreams $b_1, \ldots, b_M$, which we will name a \emph{stochastic bus of width $M$}. The $j$-th bitstream $b_j$ encodes a probability $p_j = C \cdot P([V=V_j])$.  $C$ is a bus normalization constant chosen to facilitate data encoding and processing : since $\sum\limits_j P([V=V_j]) = 1$, we have $\sum\limits_j p_j = C$. A useful choice is $C_{max} = \frac{1}{\max\limits_j P([V=V_j])}$, which allows to represent the most probable value $V_j^{max}$ by $p_j^{max} = 1$ and maximizes the p-values of other signals on the bus.

Stochastic buses can be instanciated by a set of $M$ random number generators outputting the individual bitstreams. Similarly, a set of $M$ counters can be used to recover the unnormalized probability distribution $C \cdot P(V)$.


That data representation implies that the average number of bits before observing a "1" on the $j$-th signal of the bus is $T_{avg} = \frac{1}{C \cdot P([V=V_j])}$. That number, which directly determines the number of bits necessary to get an accurate reconstruction of the distribution using counters, depends on the shape of the distribution and on the value of $C$, which is modified by the computations done on the bus and can often not be easily controlled or computed. This creates two problems. First, the number of bits necessary to reconstruct the distribution with a given desired precision can't be easily anticipated. Second, in some cases -- especially if $C$ is low -- that number may be very high, which leads to poor performance of the stochastic machine (which we call the \emph{time dilution problem}).

The first problem can be adressed by integrating the data until a given number of "1" bits have been observed on a signal, instead of during a fixed number of bits. This can easily be achieved using counters overflow. If a stochastic bitstream of width $M$ is connected to counters with a maximum value $n_{max}$, we can run the signals until one of the $M$ counters (with index ($j_{max}$) overflows. If the computation is stopped at that moment, the counter with index $j_{max}$ stores the value $n_{max}$ corresponding to the p-value $p_j^{max} = 1$, and the other counters store values $n_j$ corresponding to p-values $p_j = \frac{n_j}{n_{max}}$.

That process allows to renormalize the distribution with regard to the maximum probability value $p_j^{max}$, and to read it as a set of fixed-point numbers with precision depending on $n_{max}$. Furthermore, the index of the overflowing counter immediately gives the index of the most probable value, which implements the maximum a-posteriori estimator.

\subsubsection{Bayesian inference with stochastic bitstreams: the Bayesian machine}

\label{ssec:bm1proper}

One of the most common Bayesian computing techniques is naive Bayesian fusion \citep{Bessiere2013} : computing the posterior probability distribution on a searched variable $S$, knowing a prior distribution $P(S)$ and the conditional distributions $P(K_i|S)$ on some known variables $K_1, \ldots, K_N$. If the $K_i$ variables are conditionally independant given $S$, the inference is computed by :

\begin{equation}
P(S|K_1, \ldots, K_N) = \frac{1}{Z} P(S) \prod\limits_{i=1}^{N} P(K_i|S)
\label{eq:naivefusion}
\end{equation}
where $Z$ is a normalization constant.

\begin{figure}
\centering
\subfloat[Stochastic bitstream encoding a probability value $p = \frac{3}{8}$]{\label{sfig:bitstream}\includegraphics[width=.34\linewidth]{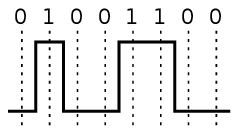}}
\hspace{.1\linewidth}
\subfloat[Computational module implementing probability product]{\label{sfig:bm1element}\includegraphics[width=.54\linewidth]{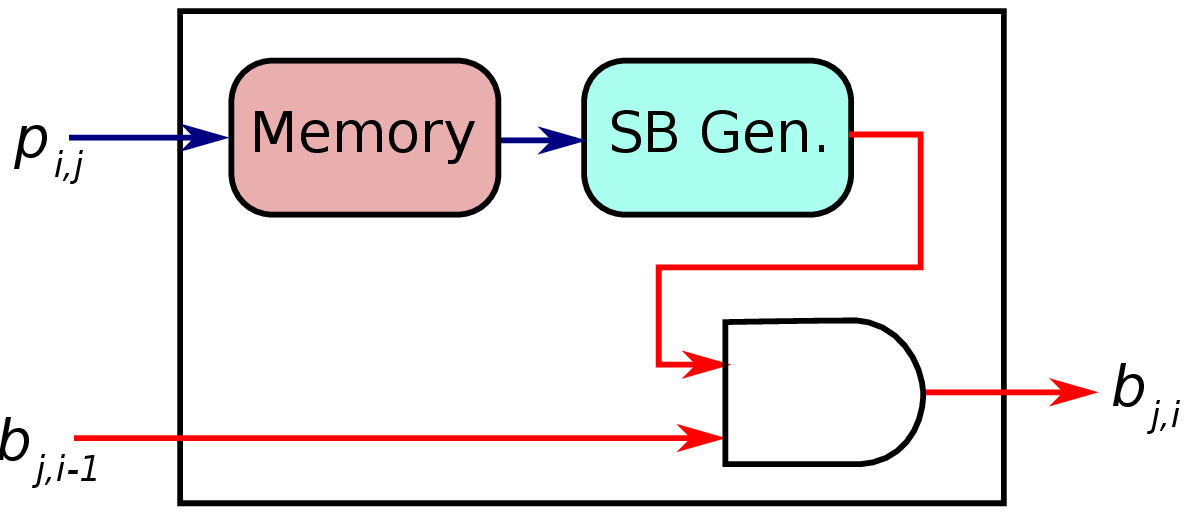}}
\caption{Computing probability products with stochastic bitstreams. Fig.~\ref{sfig:bitstream} shows 8 bits of a stochastic bitstream, with 3 bits set to 1, therefore encoding a p-value $p = \frac{3}{8}$. Fig.~\ref{sfig:bm1element} shows a computational module performing probability product as part of a Bayesian naive fusion operation: if the input signal $b_{j,i-1}$ encodes a p-value $P([S=S_j]|K_1, \ldots, K_{i-1})$, the output signal $b_{j,i}$ encodes a p-value $P([S=S_j]|K_1, \ldots, K_i) = P([S=S_j]|K_1, \ldots, K_{i-1}) \cdot C_i \cdot P(K_i|[S=S_j)$.}
\label{fig:bm1-components}
\end{figure}

This distribution can be computed using stochastic bitstreams by representing both the prior $P(S)$ and the data terms $P(K_i|S)$ with stochastic buses of width $M$, corresponding to the cardinality of $S$. After the bitstreams $b_{j,0}$ ($j \in \{1, \ldots, M\}$) encoding the prior values $P(S=S_j)$ (with a bus normalization constant $C_0$) are generated, the data terms can be integrated using simple computational modules comprised of a memory, a random generator and a logic AND gate as described in fig.~\ref{sfig:bm1element}. For each line $j \in \{1, \ldots, M\}$ in the stochastic bus and for each data term $i \in \{1, \ldots, N\}$, the memory stores the value $p_{i,j} = C_i \cdot P(K_i|S=S_j)$ (where $C_i$ is the bus normalization constant associated with data term $i$), the random generator generates a stochastic bitstream encoding probability $p_{i,j}$, and the AND gate perform the probability product between that signal and the signal $b_{j,i-1}$ from the previous element, outputting signal $b_{j,i}$.

\begin{figure}
\centering
\includegraphics[width=.99\linewidth]{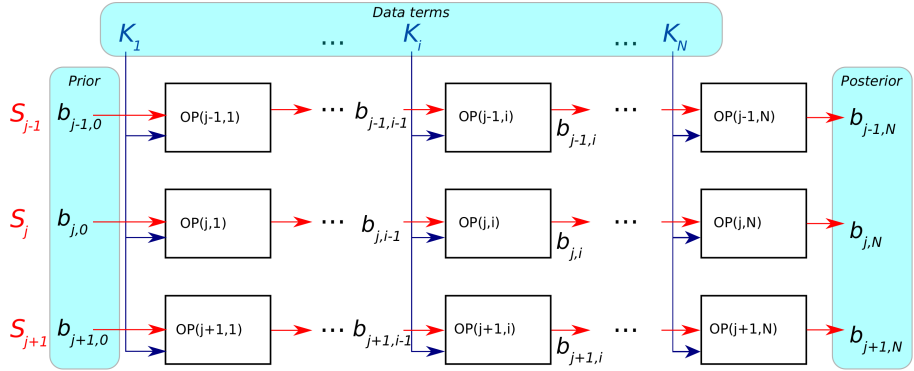}
\caption{Architecture computing naive Bayesian fusion with stochastic signals. Each OP(i,j) element is an instance of the module described in fig.~\ref{sfig:bm1element}. The leftmost input signals $b_{j,0}$ constitute a stochastic bus encoding the prior distribution $P(S)$, and the rightmost output signals $b_{j,N}$ constitute a stochastic bus encoding the posterior distribution $P(S|K_1, \ldots, K_N)$.}
\label{fig:bm1-general}
\end{figure}

The resulting architecture performs Bayesian inference using a matrix of stochastic operators, with a number of rows equal to the cardinality $M$ of variable $S$ and a number of columns equal to the number of data terms $N$ (see fig.~\ref{fig:bm1-general}). The output stochastic bus, comprised of the signals $b_{j,N}$ for $j \in \{1, \ldots, M\}$, encodes the posterior probability distribution $P(S|K_1, \ldots, K_n)$, with a bus normalization constant $C_{out} = \prod\limits_{i=0}^{N} C_i$.

\subsection{Stochastic disparity computation}
\label{ssec:bm1-disp}

\subsubsection{General description}

The architecture described in section~\ref{ssec:bm1} can be used to implement the disparity computation model described in section~\ref{sec:disparity}. The search variable is the disparity $D$, which takes values in $\llbracket 0 ; D_{max} \rrbracket$ and therefore has cardinality $D_{max}+1$, and the data terms are the three likelihood values computed from the luminance features\footnote{Color processing, with each of the three convolution filters being applied to each color layer, was also considered and experimented, but did not show significant improvement over luminance processing in the present case.} through equation~\ref{eq:lh}.

\begin{figure}
\centering
\includegraphics[width=.8\linewidth]{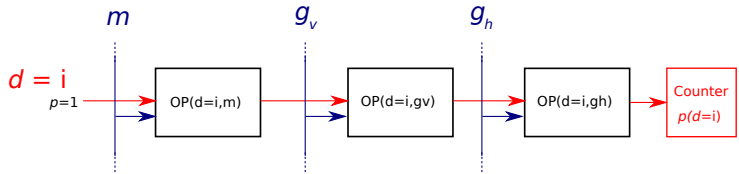}
\caption{Sequence of computational elements computing the stochastic signal corresponding to disparity value $i$. Each box contains an instance of the module described in fig.~\ref{sfig:bm1element}. A set of $D_{max}+1$ such structures allows us to compute the disparity distribution (see fig.~\ref{fig:disparity-bm1})}
\label{fig:bm1-line}
\end{figure}

We therefore use such a matrix of stochastic operators with $N = 3$ and $M = D_{max}+1$ to compute a stochastic bus representation of the posterior disparity representation. In the following, we will use a uniform disparity prior ($P([D=d]) = \frac{1}{D_{max}+1} \forall i \in \llbracket 0 ; D_{max} \rrbracket$), which can efficiently be represented by a stochastic bus with all signals constantly equal to 1 ($C_0 = D_{max}+1$). Each of the data terms are integrated as described above in section~\ref{ssec:bm1}, and the full disparity distribution for a pixel can be estimated using counters (see fig.~\ref{fig:bm1-line}). If the posterior disparity distribution is unimodal and clearly indicates a disparity value, that value can be estimated by the maximum a-posteriori estimator by simply getting the index of the first overflowing counter, as suggested in section~\ref{sssec:bus}.

\subsubsection{Processing of occlusions and low-contrast areas}

Although the previous architecture allows for efficient computation when the output distribution is unimodal and indicates a clear disparity value or a small range of values, we must adapt it to take into account some issues that arise from the fact that disparity values cannot always be computed. We will describe those problems and their consequence on the architecture, and then put forward a solution.

\begin{figure}
\centering
\includegraphics[width=.5\linewidth]{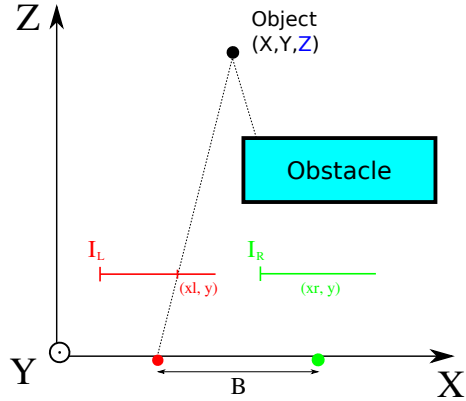}
\caption{Occlusion example: If an obstacle is positioned such as it hides a distant object from one of the two cameras, the corresponding pixels can't be matched and disparity cannot be computed.}
\label{fig:disparity-schema-occlusion}
\end{figure}

In some cases such as occlusion (see fig.~\ref{fig:disparity-schema-occlusion}), some pixels in the left image have no matching pixel in the right one and the matching costs will therefore be high for every possible disparity value. In our Bayesian model, it means the values computed by equation~\ref{eq:lh} will be small for all $d \in \llbracket 0 ; D_{max} \rrbracket$, which in our stochastic architecture translates to very low p-values for all output signals. For example, in the limit case of a pixel $(x,y)$ where the matching cost $C_f(x,y,d)$ is infinite for all disparity values $d \in \llbracket 0 ; D_{max} \rrbracket$ and for each feature $f \in \{m, g_H, g_V\}$ in equation~\ref{eq:lh}, we have $p(f^r(x-d,y)|f^l(x,y),[D(x,y)=d]) = p_0$, $\forall d \in \llbracket 0 ; D_{max} \rrbracket$, $\forall f \in \{m, g_H, g_V\}$. In our stochastic computation, all the signals in the output stochastic bus is have a p-value of $p_0^3$. This corresponds to a uniform distribution -- which is correct since no information could be inferred about the disparity value from the data -- but that distribution is encoded with a very low bus normalization constant $C=(D_{max}+1) \cdot p_0^3$, which is problematic because of the time dilution problem mentioned in section~\ref{sssec:bus}. For $p_0 = 0.02$, for example, it means that an average of one every 125000 bits will be set to 1, and the machine has to be run for an average of one million cycles simply to fill a 8-bits counter, which is very inefficient.

In some other cases, such as large uniform areas with no texture or distinctive features, the opposite problem arises: many (or possibly all) disparity values are possible match and therefore have low matching costs. The likelihood values $p(f^r(x-d,y)|f^l(x,y),[D(x,y)=d])$ then have values close to 1 for all disparity values $d \in \llbracket 0 ; D_{max} \rrbracket$, and all the the signals in the output stochastic bus will have a p-value close to 1, which encodes a high-entropy, close to uniform distribution with a high bus normalization constant. Again, this is a correct result and the high bus normalization constant means the time dilution problem does not arises; that output can efficiently be converted to numerical values or used in further stochastic computations. However, such high-entropy distributions are ill-suited to the use of the maximum a-posteriori estimator, which will return a random result among the possible disparity values.

A way to solve both those problems is to explicitly model the case where a pixel can't satisfyingly be matched, either because of occlusions or because of a lack of contrast, with an extra signal on the stochastic bus encoding a probability $p_{nomatch}$:

\begin{equation}
P(nomatch(x,y)) = p_{nm0} + (1 - p_{nm0})e^{-\frac{(g_V^l(x,y))^2}{2\sigma_{nm}^2}}
\label{eq:nomatch}
\end{equation}

The first term in the equation is a probability $p_{nm0} \gg p_0^3$ that determines the time until which an occluded pixel is detected as not matching. It should be low enough that if the pixel can be correctly matched, the stochastic signal of the corresponding disparity value $j$ has a p-value $p_j > p_{nm0}$, but high enough that if, as described above, no match is possible because of an occlusion, the ``no match'' signal fills its counter and stops the computation in a reasonable time, while detecting an absence of match.

The second term of equation~\ref{eq:nomatch} handles the poorly contrasted areas, which have been found to be characterized by low values of the vertical gradient\footnote{Note that the square of the gradient value of the left image itself is used, and not a matching cost associated to the gradient as in equation~\ref{eq:lh}. $P(nomatch(x,y))$ is therefore high if the gradient is close to zero, that is in weakly contrasted areas.} $g_V^l(x,y)$. Weakly contrasted areas will therefore have a $P(nomatch(x,y))$ value very close to 1, and the corresponding stochastic signal will very quickly fill the counter and detect an absence of match before a spurious match attributed to the behavior of the MAP estimator can be detected.

\begin{figure}
\centering
\includegraphics[width=.9\linewidth]{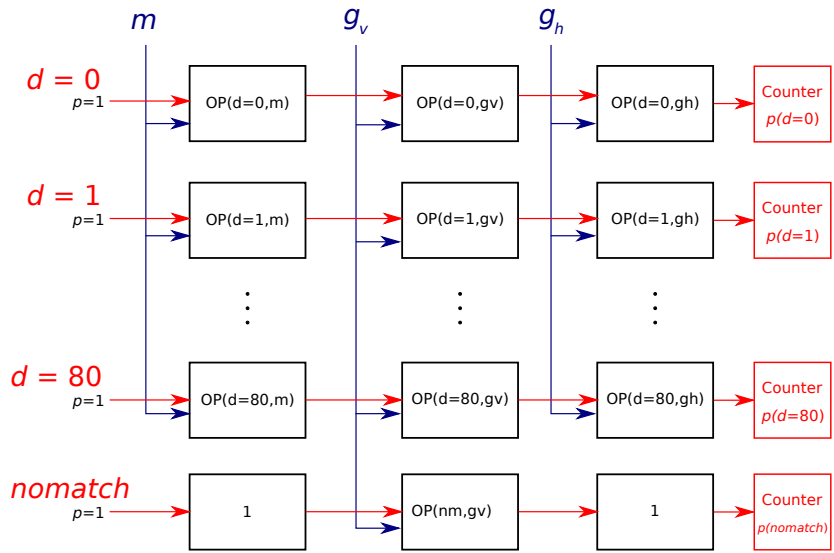}
\caption{Bayesian stochastic machine implementing the disparity computation. Each ``OP'' box contains an instance of the module described in fig.~\ref{sfig:bm1element}. Always-on signals corresponding to the uniform prior are input in the left, the feature matching likelihood values are integrated by the computational modules, and the disparity distribution is output as a stochastic bus on the right and converted to a fixed-point numeric representation by the counters. If the disparity can be computed, the counter linked to the corresponding signal will overflow first, otherwise the ``no match'' channel will overflow.}
\label{fig:disparity-bm1}
\end{figure}

The final architecture for our disparity computation stochastic machine is shown in fig.~\ref{fig:disparity-bm1}. With the extra ``no match'' signal, it has a dimension $N = 3$ and $M = D_{max}+2$.

\section{Stochastic model evaluation}
\label{sec:eval}
\subsection{Model implementation}

In order to evaluate the benefits of using a stochastic disparity computation system, we will compare two implemetations of the same Bayesian disparity algorithm described in section~\ref{ssec:model}
\begin{itemize}
\item A reference implementation performing the computation as floating point operations.
\item A simulated stochastic implementation, using pseudo-random number generators (PRNG) and bitwise boolean logic operations to simulate the behaviour of the Bayesian machine described in section~\ref{ssec:bm1-disp}.
\end{itemize}

Both implementations are programs written in C++ and run on a desktop computer equipped with an Intel Xeon E3-1271 v3 64-bit CPU. The reference implementation use FPU computations using 64 bit floating point numbers. The simulated stochastic implementation uses the Mersenne twister 19937 PRNG provided by the GNU implementation of C++11 to generate stochastic bitstreams, and the 64-bit bitwise boolean AND operation to perform probability product.

\subsection{Results}

The reference implementation ran in about 1.25 seconds per frame, which is the order of magnitude of the ``fast'' disparity algorithms from the state of the art. The simulated stochastic implementation ran in 25 to 110 seconds per frame (depending on the frame and on the size of the output counters). That low performance is due to the overhead of simulating stochastic machines using non-stochastic hardware; the performance of the stochastic system is better estimated by the number of simulated clock cycles used to compute a frame (see below).

\subsubsection{Dataset and model parameters}

We collected stereo image pairs using a PointGrey BumbleBee2 BB2-03S2C-25 wide-angle color stereoscopic camera, with focal length $f=2.5mm$, baseline distance $B=120mm$ and resolution $640 \times 480$ at 25 frames per second. The images were rectified using the Triclops proprietary PointGrey middleware. The camera was mounted on a TurtleBot 2 mobile robot base, which was manually controlled in an office environment to collect data. A total of 6301 frames were captured, corresponding to 4 minutes and 10 seconds of video.

The 24 bits color images captured were converted to 8 bits luminance images, with pixel values in $\llbracket 0 ; 255 \rrbracket$. The preprocessing described in section~\ref{ssec:model} therefore generate feature maps with pixel values in $\llbracket 0 ; 255 \rrbracket$ for the averaging filter and $\llbracket -127 ; 127 \rrbracket$ for the gradients. The $D_{max}$ value was set to 80, which corresponds to a minimum distance of 42 centimeters and was found to be adequate to the size and mobility of our robot (shorter minimum distances can easily be achieved by increasing $D_{max}$, at the price of a higher computational cost). A simple grid search performed during preliminary experiments allowed us to select good values of the other parameters, summarized in table~\ref{tab:params}.

\begin{table}[h]
\centering
\begin{tabular}{|l|c|}
\hline
Parameter & Value \\
\hline
$D_{max}$ & 80\\
$p_0$ & 0.02\\
$\sigma_{m}$ & 10\\
$\sigma_{g_V}$ & 10\\
$\sigma_{g_H}$ & 10\\
$p_{nm0}$ & 0.01\\
$\sigma_{nm}$ & 8\\
\hline
\end{tabular}
\caption{Model parameter values used for the evaluation}
\label{tab:params}
\end{table}

The feature maps were used to compute the likelihoods as described in equation~\ref{eq:lh}, and those likelihoods were used both in the reference implementation and in the simulated stochastic implementation to compute the disparity distribution.

\subsubsection{Disparity computation accuracy}
\label{sssec:results-acc}
A feature of stochastic computing using stochastic bitstreams is progressive precision: the longer the information from a bitstream is integrated, the more precisely the corresponding p-value can be estimated. In the context of the stochastic bus framework described in section~\ref{sssec:bus}, it means that precision increases with the size of the counters used to estimate the distribution: the higher the counters' maximum value, the closer to the reference implementation the resulting distribution is expected to be. We therefore used the simulated stochastic implementation with variable counter sizes to quantify that phenomenon.

The stochastic disparity processor described in section~\ref{ssec:bm1-disp} performs two functions: detecting the ``no match'' pixels, and computing the disparity distribution on matched pixels. The performance of the ``no match'' pixels discrimination task can be assessed by computing the F1-score between the set of pixels identified as ``no match '' by the reference implementation and the simulated stochastic implementation. The performance of the disparity distribution computation task can be assessed by measuring the RMS error between the distributions estimated from the simulated stochastic implementation and the reference implementation\footnote{Using the KL-divergence has also been considered, but proved problematic because of the frequent occurence of 0 as a p-value in the distributions estimated from the simulated stochastic implementation.}.

\begin{figure}
\centering
\centering
\subfloat[Global RMS error (for matched pixels) and F1-score (for non-match detection)]{\label{fig:cbm1-accuracy}\includegraphics[width=.48\linewidth]{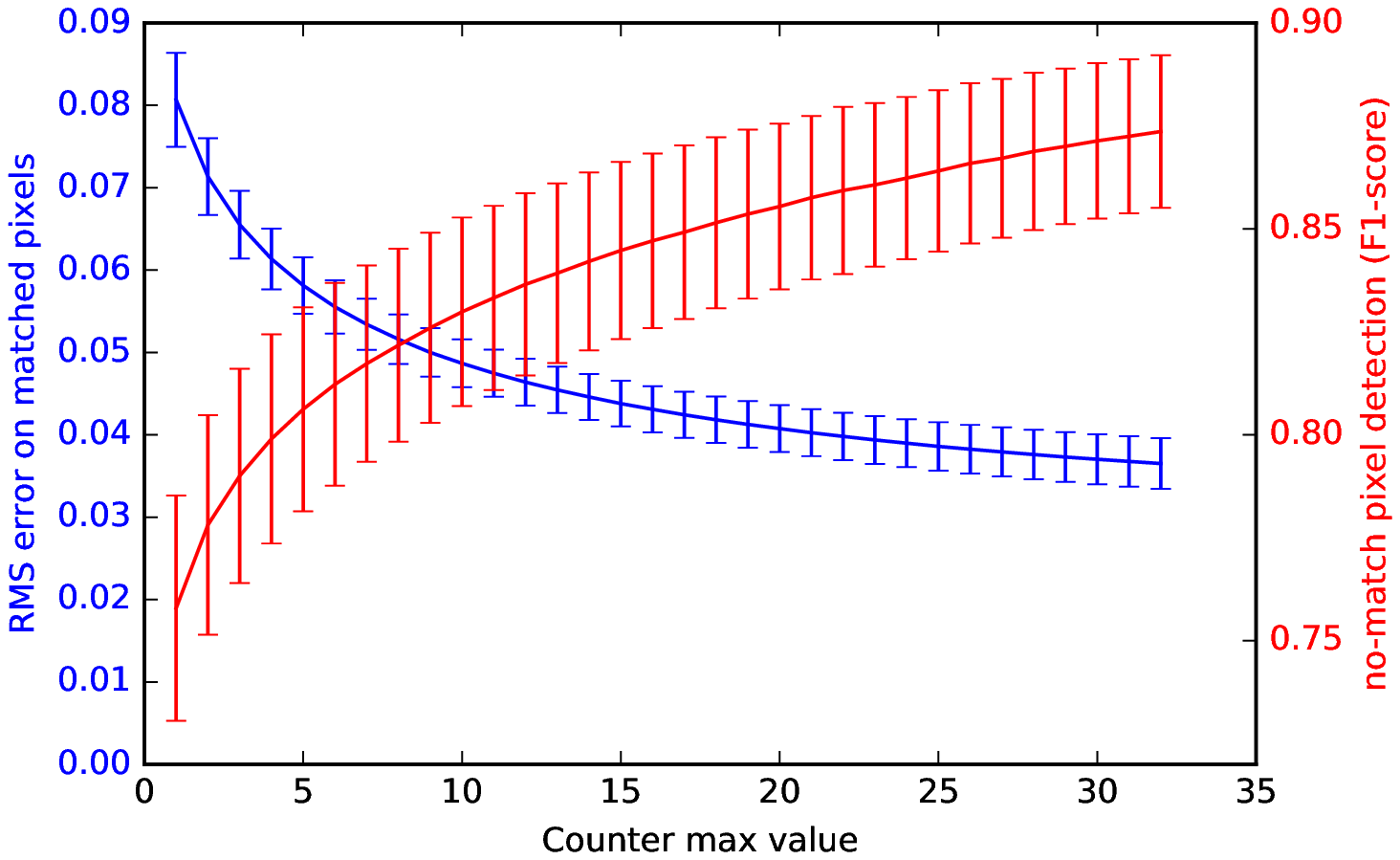}}
\hspace{.03\linewidth}
\subfloat[Average run time per pixel]{\label{fig:cbm1-speed}\includegraphics[width=.48\linewidth]{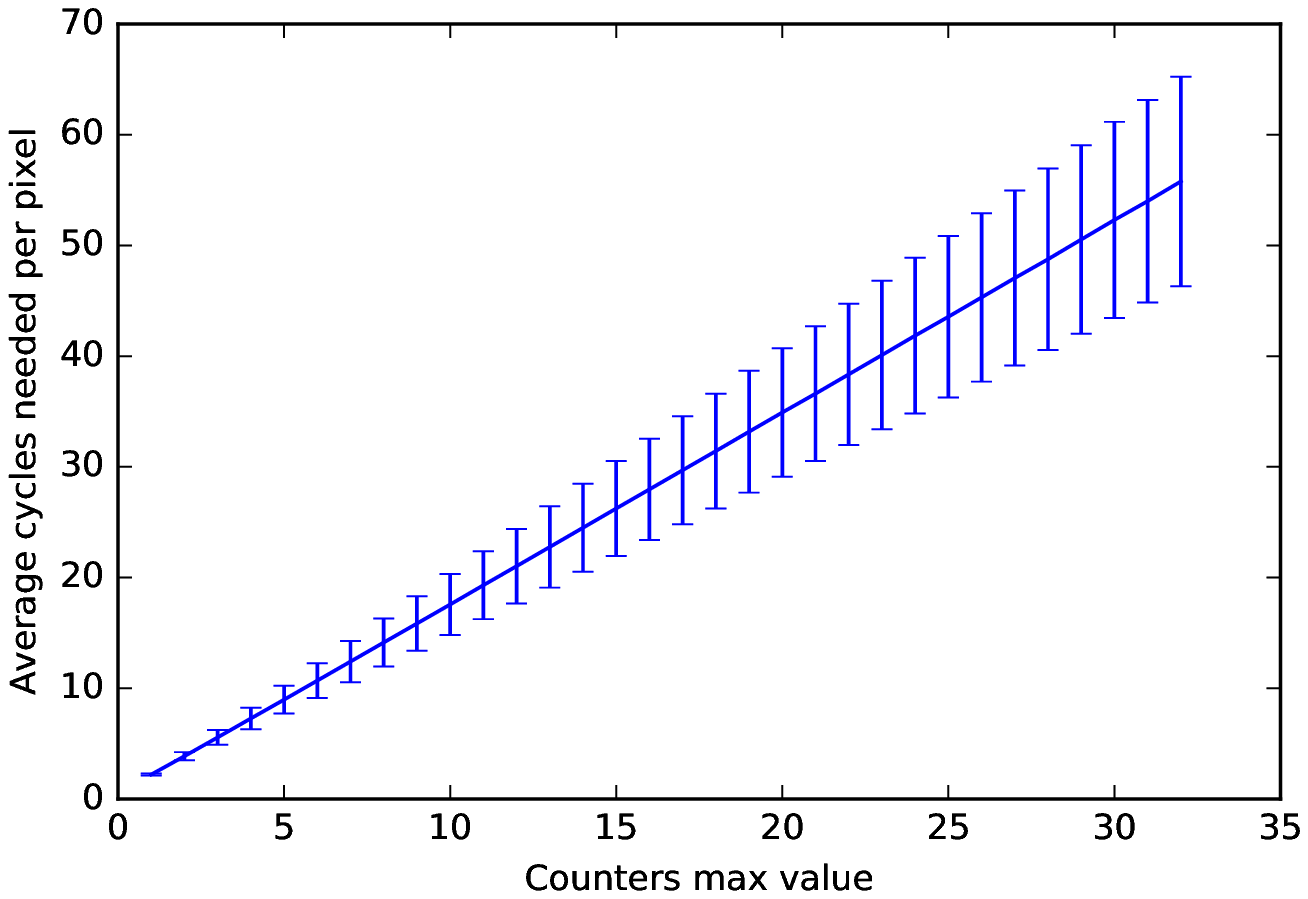}}
\caption{Performance of the simulated stochastic disparity processor. Fig.~\ref{fig:cbm1-accuracy} shows that the detection of ``no match'' pixels and the quality of the reconstructed distribution exponentially improve with counter max value. Fig.~\ref{fig:cbm1-speed} depicts the linear relationship between counter max value and simulated run time (in number of simulated clock cycles to compute the disparity distribution for one pixel). The data shows the mean and standard deviation on a subset of 126 images obtained by regular sampling of the full dataset.}
\label{fig:cbm1-results}
\end{figure}

Figure~\ref{fig:cbm1-accuracy} shows that indeed, the F1-score exponentially grows closer to 1 and the RMS error exponentially decreases with the counter max value. For example, with 16-bits counters the RMS error is below $0.05$ and the F1-score above 80\%.

Figure~\ref{fig:cbm1-speed} shows the relationship between the counter max value and the average time (in number of clock cycles) the simulated stochastic machine has to run before filling a counter. As expected, it grows linearly with maximum value of the counter: as all signals are uncorrelated, each extra ``1'' required to fill the counter generates a constant overhead.

\begin{figure}
\centering
\subfloat[Left image]{\label{fig:base-image}\includegraphics[width=.45\linewidth]{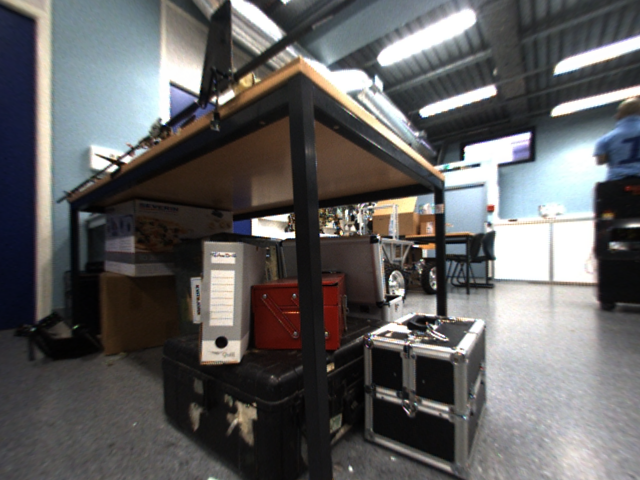}}
\hspace{.05\linewidth}
\subfloat[Reference disparity image (64 bits floating point computation)]{\label{fig:image-disp-ref}\includegraphics[width=.4\linewidth]{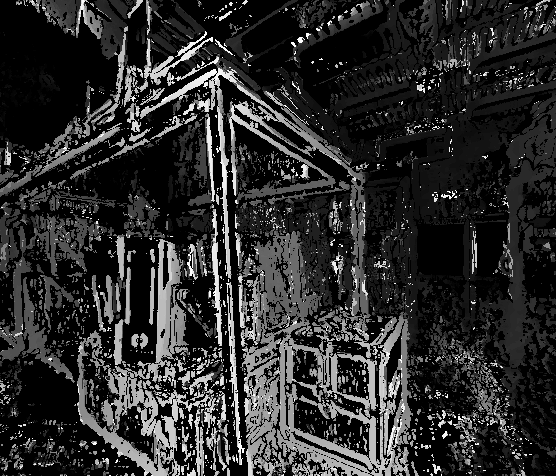}}\\
\vspace{.05\linewidth}
\subfloat[Stochastic computation, counter max value 1]{\label{fig:image-disp-1}\includegraphics[width=.45\linewidth]{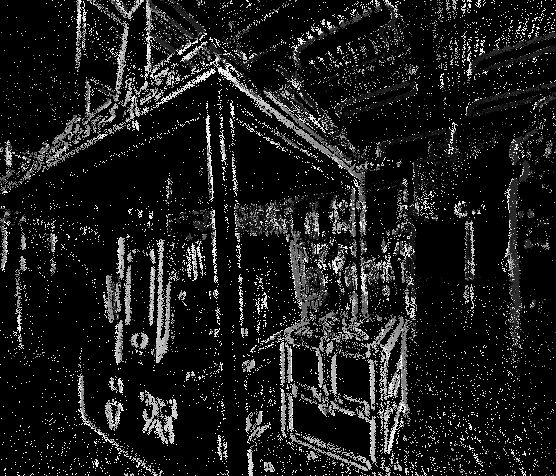}}
\hspace{.05\linewidth}
\subfloat[Stochastic computation, counter max value 16]{\label{fig:image-disp-16}\includegraphics[width=.45\linewidth]{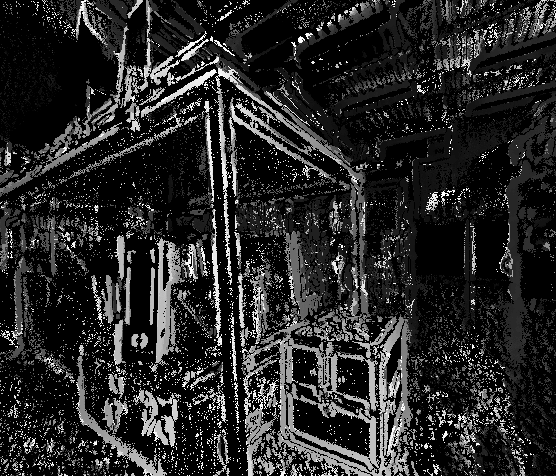}}
\caption{Disparity images obtained by applying the MAP estimator to the output distribution for an exemple frame. Fig.~\ref{fig:base-image} is the rectified frame from the left camera. Fig.~\ref{fig:image-disp-ref} is the reference image obtained by classic 64 bits floating point computation. Fig.~\ref{fig:image-disp-1} and \ref{fig:image-disp-16} show the images obtained by the simulated stochastic implementation, respectively with max counter values 1 and 16. Black pixels are ``no match'' pixels; white pixels have disparity to $d = D_{max} = 80$, and gray pixels have a luminance proportional to their disparity value in $\llbracket 0 ; D_{max} \rrbracket$}
\label{fig:dispimage}
\end{figure}

Fig.~\ref{fig:dispimage} shows an example of reconstructed disparity image with the reference implementation, and with the simulated stochastic implementation using two maximum counter values, 1 and 16. The disparity image from the stochastic system with the larger counters is visually close to the reference. The image obtained using 1-bit counters, while clearly noisier and lower quality, still correctly describes the general tridimensional structure of the scene and could possibly be used to drive a robust robot control system. According to the data from fig.~\ref{fig:cbm1-speed}, the stochastic computation of the disparity distribution requires $2.21 \pm 0.09$ clock cycles per pixel for 1-bit counters and $27.97 \pm 4.58$ clock cycles per pixel for counters with a maximum value of 16. Both those values compare favorably to the floating point computations performing to the same operations, which requires at least $81 \times 3$ floating point number products and a maximum search on a 81-value vector.

\section{Discussion: speed, energy and hardware implementation considerations}
\label{sec:discu}
The above results show that our stochastic computational system can successfully implement a Bayesian binocular disparity algorithm and compute full disparity distribution with good accuracy, using stochastic bitstreams and a reduced number of computation cycles.

However, the stochastic bitstream-based computational system described in section~\ref{ssec:bm1} supposes the use of fast, efficient sources of stochastic signals, that could be integrated at a large scale in a hardware component, jointly with AND gates, memories and counters, to implement the architecture seen in section~\ref{ssec:bm1proper}. In this paper, we used a simulated implementation using Mersenne twister PRNGs to evaluate the potential of this approach in the absence of such components. But recent advances in new nanodevices based on spintronics, such as the superparamagnetic tunnel junction (SMTJ) \citep{Locatelli2014,Locatelli2015}, bear the promise that such generators could be available in the short or medium term. Experimental SMTJ devices have been shown to be able to generate high-quality stochastic bitstreams at a frequency of 500MHz with a very low power consumption of \SI{50}{\micro\watt}. Those components can be built with CMOS technology using an area equivalent to 12 bytes of SRAM \citep{Querlioz-annual-meeting}, making them suitable to large scale integration with the other components needed to build the stochastic machines described above.

Using those figures as guidelines, we can compute the order of magnitude of the speed and power consumption of the disparity computation systen described in section~\ref{ssec:bm1-disp} and evaluated in simulation in section~\ref{sec:eval}. The system requires 246 random signal generators, which would have a total power consumption of \SI{12.3}{\milli\watt}. Using counters with a maximum value of 16, which has been shown in section~\ref{sssec:results-acc} to be an adequate tradeoff between speed and accuracy, we need an average of 27.97 clock cycles per pixel, with $(640 - 4 - 80) \times (480 -4)$\footnote{Each dimension of the original $640 \times 480$ images is reduced by 4 pixels by the prefiltering as seen in section~\ref{ssec:model}, and the horizontal dimension is further reduced by $D_{max}$ since the distribution can't be computed for the $D_{max}$ first pixels of each row as shown by equation~\ref{eq:cost}.}, which represents an average total of 7402428.32 clock cycles per image. At a frequency of 500MHz, the architecture would therefore be able to process about 67.5 image pairs per second. As our system processes data for each pixel independantly, computation time and power consumption are expected to grow linearly with image width and height.

Those computations are only rough estimations; more specifically the energy consumption computation ignores the energy cost of the AND gates, memories and counters also necessary to implement the circuit, and the performance does not take into account the overhead induced by reinitializing the machine between each pixel (resetting the counters and loading the data memories). Our Bayesian algorithm also makes use of preprocessed images using spatial filters; the cost (both computational and energetic) of that preprocessing should be taken into account into any global evaluation of the system. But many methods exist to perform such spatial filtering (using general-purpose CPUs, GPUs, FPGAs, dedicated hardware, etc.) with various cost, performance and energy-efficiency characteristics; further work will explore ways through which such filtering could be done using stochastic computations. Similarly, the cost computation step is currently performed using classic floating-point computation, the opportunity to use stochastic computations instead is currently being studied.

On the other hand, those computations are assumed to be performed sequentially for each pixel on a unique instance of the systems described in section~\ref{ssec:bm1-disp}, using only 246 of the computing modules described in fig.~\ref{sfig:bm1element}. But our Bayesian machine architecture is parallel by design, and a higher number of those modules would allow for parallel processing of many pixels and increased performance, at the cost of higher circuit size and energy consumption.




\section{Conclusion}
\label{sec:ccl}

We have put forward an architecture to compute a class of Bayesian inference problems with probabilistic hardware using stochastic bitstreams, and evaluated that system in simulation on the example of binocular disparity computation, demonstrating high performance and energy-efficiency. Although the work described in this paper uses simulations of hypothetical stochastic machines using experimental hardware devices and can therefore only offer rough estimations of the performance of those systems, it is our belief that those results clearly highlight the potential of Bayesian computation using stochastic bitstreams for sensorimotor processing, especially in applications with tight constraints on computational and energy resources such as mobile robotics, embedded systems or distributed sensors.

The proposed architecture allows to solve many sensory fusion and processing problems, yielding full distributions expressed as bus of stochastic bitstreams, with a low power consumption and reduced computational resources. The parallel and distributed nature of our architecture could allow to easily address a variety of different problems using the same arrays of generic components, in a way similar to FPGAs. Furthermore, the progressive precision of the stochastic bitstream data representation allows to easily adjust the speed/accuracy or power/accuracy tradeoffs by changing the signal integration time (determined by the counters maximum values), making it possible, for example, to maintain degraded operation with lower accuracy in low energy conditions.

Future work will entail continued collaboration with projects partner to physically instantiate the system described and simulated inm the present work. A partial implementation of our Bayesian Machine infrastructure using FPGA systems has been demonstrated \citep{Coninx2016icrc}, and further research will also integrate the technology developed by teams working on stochastic signal generator devices. Efforts will also be dedicated to extending the breadth of the computations implemented by stochastic bitstream based systems, combining the disparity computation to other sensory computations (such as optical flow) to create an occupancy map, using and extending existing Bayesian spatial cognition algorithms \citep{Elfes1989,Thrun2003,Coue2006} which could then be used for obstacle avoidance and robot navigation, paving the way to a completely stochastic robot sensorimotor controller.


\section*{Acknowledgements}
This work was performed within the EU Future and Emerging Technologies \href{https://www.bambi-fet.eu/}{BAMBI project} [FP7-ICT-2013-
C, project number 618024]. It was also partly supported by \href{http://www.smart-labex.fr/}{ANR Labex SMART} [ANR-11-LABX-65].

\section*{References}

\bibliography{library}

\begin{thebibliography}{33}
\expandafter\ifx\csname natexlab\endcsname\relax\def\natexlab#1{#1}\fi
\providecommand{\url}[1]{\texttt{#1}}
\providecommand{\href}[2]{#2}
\providecommand{\path}[1]{#1}
\providecommand{\DOIprefix}{doi:}
\providecommand{\ArXivprefix}{arXiv:}
\providecommand{\URLprefix}{URL: }
\providecommand{\Pubmedprefix}{pmid:}
\providecommand{\doi}[1]{\href{http://dx.doi.org/#1}{\path{#1}}}
\providecommand{\Pubmed}[1]{\href{pmid:#1}{\path{#1}}}
\providecommand{\bibinfo}[2]{#2}
\ifx\xfnm\relax \def\xfnm[#1]{\unskip,\space#1}\fi
\bibitem[{Geiger et~al.(2012)Geiger, Lenz, and Urtasun}]{Geiger2012}
\bibinfo{author}{A.~Geiger}, \bibinfo{author}{P.~Lenz},
  \bibinfo{author}{R.~Urtasun},
\newblock \bibinfo{title}{{Are we ready for Autonomous Driving? The
  $\backslash$textsc{\{}KITTI{\}} Vision Benchmark Suite}},
\newblock \bibinfo{journal}{Computer Vision and Pattern Recognition)}
  (\bibinfo{year}{2012}).
\bibitem[{Scharstein and Szeliski(2002)}]{Scharstein2002}
\bibinfo{author}{D.~Scharstein}, \bibinfo{author}{R.~Szeliski},
\newblock \bibinfo{title}{{A taxonomy and evaluation of dense two-frame stereo
  correspondence algorithms}},
\newblock \bibinfo{journal}{International Journal of Computer Vision}
  \bibinfo{volume}{47} (\bibinfo{year}{2002}) \bibinfo{pages}{7--42}.
\bibitem[{Lazaros et~al.(2008)Lazaros, Sirakoulis, and
  Gasteratos}]{Lazaros2008}
\bibinfo{author}{N.~Lazaros}, \bibinfo{author}{G.~C. Sirakoulis},
  \bibinfo{author}{A.~Gasteratos},
\newblock \bibinfo{title}{{Review of Stereo Vision Algorithms: From Software to
  Hardware}},
\newblock \bibinfo{journal}{International Journal of Optomechatronics}
  \bibinfo{volume}{2} (\bibinfo{year}{2008}) \bibinfo{pages}{435--462}.
\bibitem[{Geiger et~al.(2013)Geiger, Lenz, Stiller, and Urtasun}]{Geiger2013}
\bibinfo{author}{a.~Geiger}, \bibinfo{author}{P.~Lenz},
  \bibinfo{author}{C.~Stiller}, \bibinfo{author}{R.~Urtasun},
\newblock \bibinfo{title}{{Vision meets robotics: The KITTI dataset}},
\newblock \bibinfo{journal}{The International Journal of Robotics Research}
  \bibinfo{volume}{32} (\bibinfo{year}{2013}) \bibinfo{pages}{1231--1237}.
\bibitem[{Scharstein et~al.(2014)Scharstein, Hirschm{\"{u}}ller, Kitajima,
  Krathwohl, Ne{\v{s}}i{\'{c}}, Wang, and Westling}]{Scharstein2014}
\bibinfo{author}{D.~Scharstein}, \bibinfo{author}{H.~Hirschm{\"{u}}ller},
  \bibinfo{author}{Y.~Kitajima}, \bibinfo{author}{G.~Krathwohl},
  \bibinfo{author}{N.~Ne{\v{s}}i{\'{c}}}, \bibinfo{author}{X.~Wang},
  \bibinfo{author}{P.~Westling},
\newblock \bibinfo{title}{{High-resolution stereo datasets with
  subpixel-accurate ground truth}},
\newblock \bibinfo{journal}{Lecture Notes in Computer Science (including
  subseries Lecture Notes in Artificial Intelligence and Lecture Notes in
  Bioinformatics)} \bibinfo{volume}{8753} (\bibinfo{year}{2014})
  \bibinfo{pages}{31--42}.
\bibitem[{Belhumeur(1996)}]{Belhumeur1996}
\bibinfo{author}{P.~N. Belhumeur},
\newblock \bibinfo{title}{{A Bayesian Approach to Binocular Stereopsis}},
\newblock \bibinfo{journal}{International Journal of Computer Vision}
  \bibinfo{volume}{19} (\bibinfo{year}{1996}) \bibinfo{pages}{237--260}.
\bibitem[{Su et~al.(2012)Su, Bovik, and Cormack}]{Su2012}
\bibinfo{author}{C.~C. Su}, \bibinfo{author}{A.~C. Bovik},
  \bibinfo{author}{L.~K. Cormack},
\newblock \bibinfo{title}{{Statistical model of color and disparity with
  application to Bayesian stereopsis}},
\newblock \bibinfo{journal}{Proceedings of the IEEE Southwest Symposium on
  Image Analysis and Interpretation}  (\bibinfo{year}{2012})
  \bibinfo{pages}{169--172}.
\bibitem[{Coue et~al.(2006)Coue, Pradalier, Laugier, Fraichard, and
  Bessiere}]{Coue2006}
\bibinfo{author}{C.~Coue}, \bibinfo{author}{C.~Pradalier},
  \bibinfo{author}{C.~Laugier}, \bibinfo{author}{T.~Fraichard},
  \bibinfo{author}{P.~Bessiere},
\newblock \bibinfo{title}{{Bayesian Occupancy Filtering for Multitarget
  Tracking: An Automotive Application}},
\newblock \bibinfo{journal}{The International Journal of Robotics Research}
  \bibinfo{volume}{25} (\bibinfo{year}{2006}) \bibinfo{pages}{19--30}.
\bibitem[{Thrun et~al.(2005)Thrun, Burgard, and Fox}]{Thrun2005}
\bibinfo{author}{S.~Thrun}, \bibinfo{author}{W.~Burgard},
  \bibinfo{author}{D.~Fox}, \bibinfo{title}{{Probabilistic Robotics}},
  \bibinfo{publisher}{MIT Press}, \bibinfo{year}{2005}.
\bibitem[{Lebeltel(2006)}]{Lebeltel2006}
\bibinfo{author}{O.~Lebeltel}, \bibinfo{title}{{Programmation Bay{\'{e}}sienne
  des Robots}}, Ph.D. thesis, Universit{\'{e}} de Grenoble,
  \bibinfo{year}{2006}.
\bibitem[{Bessi{\`{e}}re et~al.(2008)Bessi{\`{e}}re, Laugier, and
  Siegwart}]{Bessiere2008}
\bibinfo{author}{P.~Bessi{\`{e}}re}, \bibinfo{author}{C.~Laugier},
  \bibinfo{author}{R.~Siegwart}, \bibinfo{title}{{Probabilistic Reasoning and
  Decision Making in Sensory-Motor Systems}}, volume~\bibinfo{volume}{46} of
  \textit{\bibinfo{series}{Springer Tracts in Advanced Robotics}},
  \bibinfo{publisher}{Springer Berlin Heidelberg}, \bibinfo{address}{Berlin,
  Heidelberg}, \bibinfo{year}{2008}. \URLprefix
  \url{http://link.springer.com/10.1007/978-3-540-79007-5}.
  \DOIprefix\doi{10.1007/978-3-540-79007-5}.
  \href{http://arxiv.org/abs/arXiv:1011.1669v3}{\tt arXiv:arXiv:1011.1669v3}.
\bibitem[{Alves et~al.(2015)Alves, Ferreira, Lobo, and Dias}]{Alves2015}
\bibinfo{author}{J.~D. Alves}, \bibinfo{author}{J.~F. Ferreira},
  \bibinfo{author}{J.~Lobo}, \bibinfo{author}{J.~Dias},
\newblock \bibinfo{title}{{Brief Survey on Computational Solutions for Bayesian
  Inference}},
\newblock in: \bibinfo{booktitle}{Workshop on Unconventional computing for
  Bayesian inference at IROS2015}, \bibinfo{address}{Hamburg},
  \bibinfo{year}{2015}.
\bibitem[{{Von Neumann}(1956)}]{VonNeumann1956}
\bibinfo{author}{J.~{Von Neumann}}, \bibinfo{title}{{Probabilistic logics and
  the synthesis of reliable organisms from unreliable components}},
  \bibinfo{year}{1956}. \URLprefix
  \url{http://books.google.com/books?hl=en{\&}lr={\&}id=QaruU73YWGkC{\&}oi=fnd{\&}pg=PA110{\&}dq=PROBABILISTIC+LOGICS+AND+THE+SYNTHESIS+OP+RELIABLE.+ORGANISMS+PROM+UNRELIABLE+COMPONENTS{\&}ots=AdOY3cy2Nu{\&}sig=i80DJDGUrK51AETEzojzVtx5LwM}.
  \DOIprefix\doi{10.1128/AEM.00314-09}.
\bibitem[{Gaines(1969)}]{Gaines1969}
\bibinfo{author}{B.~Gaines},
\newblock \bibinfo{title}{{Stochastic computing systems}},
\newblock \bibinfo{journal}{Advances in information systems science}
  (\bibinfo{year}{1969}) \bibinfo{pages}{37--172}.
\bibitem[{Vigoda(2003)}]{Vigoda2003}
\bibinfo{author}{B.~Vigoda}, \bibinfo{title}{{Analog Logic : Continuous-Time
  Analog Circuits for Statistical Signal Processing}}, Ph.D. thesis,
  Massachusetts Institute of Technology, \bibinfo{year}{2003}. \URLprefix
  \url{http://pubs.media.mit.edu/pubs/papers/03.07.vigoda.pdf}.
\bibitem[{Mansinghka(2009)}]{Mansinghka2009}
\bibinfo{author}{V.~Mansinghka}, \bibinfo{title}{{Natively Probabilistic
  Computation}}, Ph.D. thesis, Massachusetts Institute of Technology,
  \bibinfo{year}{2009}. \URLprefix
  \url{http://dspace.mit.edu/handle/1721.1/47892}.
\bibitem[{Jonas et~al.(2014)Jonas, Tenenbaum, and Wilson}]{Jonas2014}
\bibinfo{author}{E.~Jonas}, \bibinfo{author}{J.~B. Tenenbaum},
  \bibinfo{author}{M.~a. Wilson}, \bibinfo{title}{{Stochastic Architectures for
  Probabilistic Computation by}}, Ph.D. thesis, Massachssets Institute of
  Technology, \bibinfo{year}{2014}.
\bibitem[{Khasanvis et~al.(2015)Khasanvis, Li, Rahman, Salehi-Fashami, Biswas,
  Atulasimha, Bandyopadhyay, and Moritz}]{Khasanvis2015}
\bibinfo{author}{S.~Khasanvis}, \bibinfo{author}{M.~Li},
  \bibinfo{author}{M.~Rahman}, \bibinfo{author}{M.~Salehi-Fashami},
  \bibinfo{author}{A.~K. Biswas}, \bibinfo{author}{J.~Atulasimha},
  \bibinfo{author}{S.~Bandyopadhyay}, \bibinfo{author}{C.~A. Moritz},
\newblock \bibinfo{title}{{Self-Similar Magneto-Electric Nanocircuit Technology
  for Probabilistic Inference Engines}},
\newblock \bibinfo{journal}{IEEE Transactions on Nanotechnology}
  \bibinfo{volume}{14} (\bibinfo{year}{2015}) \bibinfo{pages}{980--991}.
\bibitem[{Ferreira et~al.(2015)Ferreira, Lanillos, and Dias}]{Ferreira2015}
\bibinfo{author}{J.~F. Ferreira}, \bibinfo{author}{P.~Lanillos},
  \bibinfo{author}{J.~Dias},
\newblock \bibinfo{title}{{Fast Exact Bayesian Inference for High-Dimensional
  Models}},
\newblock in: \bibinfo{booktitle}{Workshop on Unconventional computing for
  Bayesian inference (UCBI), IEEE/RSJ International Conference on Intelligent
  Robots and Systems (IROS)}, \bibinfo{year}{2015}. \URLprefix
  \url{http://hgpu.org/?p=14611}.
\bibitem[{Thakur et~al.(2016)Thakur, Afshar, Wang, Hamilton, Tapson, and van
  Schaik}]{Thakur2016}
\bibinfo{author}{C.~S. Thakur}, \bibinfo{author}{S.~Afshar},
  \bibinfo{author}{R.~M. Wang}, \bibinfo{author}{T.~J. Hamilton},
  \bibinfo{author}{J.~Tapson}, \bibinfo{author}{A.~van Schaik},
\newblock \bibinfo{title}{{Bayesian Estimation and Inference using Stochastic
  Hardware}},
\newblock \bibinfo{journal}{Frontiers in Neuroscience} \bibinfo{volume}{10}
  (\bibinfo{year}{2016}) \bibinfo{pages}{1--28}.
\bibitem[{Friedman et~al.(2016)Friedman, Calvet, Bessiere, Droulez, and
  Querlioz}]{Friedman2016}
\bibinfo{author}{J.~S. Friedman}, \bibinfo{author}{L.~E. Calvet},
  \bibinfo{author}{P.~Bessiere}, \bibinfo{author}{J.~Droulez},
  \bibinfo{author}{D.~Querlioz},
\newblock \bibinfo{title}{{Bayesian Inference With Muller C-Elements}},
\newblock \bibinfo{journal}{IEEE Transactions on Circuits and Systems I:
  Regular Papers} \bibinfo{volume}{In Press} (\bibinfo{year}{2016})
  \bibinfo{pages}{1--10}.
\bibitem[{Faix et~al.(2015)Faix, Lobo, Laurent, Vaufreydaz, and
  Mazer}]{Faix2015}
\bibinfo{author}{M.~Faix}, \bibinfo{author}{J.~Lobo},
  \bibinfo{author}{R.~Laurent}, \bibinfo{author}{D.~Vaufreydaz},
  \bibinfo{author}{E.~Mazer},
\newblock \bibinfo{title}{{Stochastic Bayesian Computation for Autonomous Robot
  Sensorimotor Systems}},
\newblock in: \bibinfo{booktitle}{Proceedings of the IROS2015 workshop on
  Unconventional computing for Bayesian inference}, \bibinfo{year}{2015}, pp.
  \bibinfo{pages}{27--32}.
\bibitem[{Scharstein(1994)}]{Scharstein1994}
\bibinfo{author}{D.~Scharstein},
\newblock \bibinfo{title}{{Matching images by comparing their gradient
  fields}},
\newblock \bibinfo{journal}{Proceedings of 12th International Conference on
  Pattern Recognition} \bibinfo{volume}{1} (\bibinfo{year}{1994})
  \bibinfo{pages}{4--7}.
\bibitem[{Jones and Malik(1992)}]{Jones1992}
\bibinfo{author}{D.~G. Jones}, \bibinfo{author}{J.~Malik},
\newblock \bibinfo{title}{{A Computational framework for determining stereo
  correspondence from a set of linear spatial filters}},
\newblock \bibinfo{journal}{Image and Vision Computing} \bibinfo{volume}{10}
  (\bibinfo{year}{1992}) \bibinfo{pages}{699----708}.
\bibitem[{{\v{Z}}bontar and LeCun(2014)}]{Zbontar2014}
\bibinfo{author}{J.~{\v{Z}}bontar}, \bibinfo{author}{Y.~LeCun},
\newblock \bibinfo{title}{{Computing the Stereo Matching Cost with a
  Convolutional Neural Network}},
\newblock \bibinfo{journal}{arXiv preprint arXiv:1409.4326}
  (\bibinfo{year}{2014}).
\bibitem[{Mayer et~al.(2015)Mayer, Ilg, H{\"{a}}usser, Fischer, Cremers,
  Dosovitskiy, and Brox}]{Mayer2015}
\bibinfo{author}{N.~Mayer}, \bibinfo{author}{E.~Ilg},
  \bibinfo{author}{P.~H{\"{a}}usser}, \bibinfo{author}{P.~Fischer},
  \bibinfo{author}{D.~Cremers}, \bibinfo{author}{A.~Dosovitskiy},
  \bibinfo{author}{T.~Brox}, \bibinfo{title}{{A Large Dataset to Train
  Convolutional Networks for Disparity, Optical Flow, and Scene Flow
  Estimation}}, \bibinfo{type}{Technical Report}, arXiv preprint 1512.02134,
  \bibinfo{year}{2015}. \URLprefix \url{http://arxiv.org/abs/1512.02134}.
  \href{http://arxiv.org/abs/1512.02134}{\tt arXiv:1512.02134}.
\bibitem[{Bessi{\`{e}}re et~al.(2013)Bessi{\`{e}}re, Ahuactzin, Mekhnacha, and
  Mazer}]{Bessiere2013}
\bibinfo{author}{P.~Bessi{\`{e}}re}, \bibinfo{author}{J.-M. Ahuactzin},
  \bibinfo{author}{K.~Mekhnacha}, \bibinfo{author}{E.~Mazer},
  \bibinfo{title}{{Bayesian Programming}}, \bibinfo{publisher}{Chapman and
  Hall/CRC}, \bibinfo{year}{2013}. \URLprefix
  \url{ftp://ftp.inrialpes.fr/pub/emotion/bayesian.../Bayesian-Programming.pdf}.
\bibitem[{Locatelli et~al.(2014)Locatelli, Mizrahi, Accioly, Querlioz, Kim,
  Cros, and Grollier}]{Locatelli2014}
\bibinfo{author}{N.~Locatelli}, \bibinfo{author}{A.~Mizrahi},
  \bibinfo{author}{A.~Accioly}, \bibinfo{author}{D.~Querlioz},
  \bibinfo{author}{J.-V. Kim}, \bibinfo{author}{V.~Cros},
  \bibinfo{author}{J.~Grollier},
\newblock \bibinfo{title}{{Spin torque nanodevices for bio-inspired
  computing}},
\newblock in: \bibinfo{booktitle}{2014 14th International Workshop on Cellular
  Nanoscale Networks and their Applications (CNNA)},
  volume~\bibinfo{volume}{1}, \bibinfo{publisher}{IEEE}, \bibinfo{year}{2014},
  pp. \bibinfo{pages}{1--2}. \URLprefix
  \url{http://dx.doi.org/10.1038/ncomms1006
  http://ieeexplore.ieee.org/lpdocs/epic03/wrapper.htm?arnumber=6888659}.
  \DOIprefix\doi{10.1109/CNNA.2014.6888659}.
  \href{http://arxiv.org/abs/arXiv:1001.4933v1}{\tt arXiv:arXiv:1001.4933v1}.
\bibitem[{Locatelli et~al.(2015)Locatelli, Vincent, Mizrahi, Friedman,
  Vodenicarevic, Kim, Klein, Zhao, Grollier, and Querlioz}]{Locatelli2015}
\bibinfo{author}{N.~Locatelli}, \bibinfo{author}{A.~F. Vincent},
  \bibinfo{author}{A.~Mizrahi}, \bibinfo{author}{J.~S. Friedman},
  \bibinfo{author}{D.~Vodenicarevic}, \bibinfo{author}{J.-V. Kim},
  \bibinfo{author}{J.-O. Klein}, \bibinfo{author}{W.~Zhao},
  \bibinfo{author}{J.~Grollier}, \bibinfo{author}{D.~Querlioz},
\newblock \bibinfo{title}{{Spintronic Devices as Key Elements for
  Energy-Efficient Neuroinspired Architectures}},
\newblock \bibinfo{journal}{Proceedings of the 2015 Design, Automation {\&}
  Test in Europe Conference {\&} Exhibition} \bibinfo{volume}{1}
  (\bibinfo{year}{2015}) \bibinfo{pages}{994--999}.
\bibitem[{Querlioz(2016)}]{Querlioz-annual-meeting}
\bibinfo{author}{D.~Querlioz},
\newblock \bibinfo{title}{{Review of IEF's work - Modelling of
  superparamagnetic MTJs}},
\newblock in: \bibinfo{booktitle}{BAMBI-FET second year annual meeting},
  \bibinfo{address}{Paris}, \bibinfo{year}{2016}.
\bibitem[{Coninx et~al.(2016)Coninx, Laurent, Aslam, Bessi{\`{e}}re, Lobo,
  Mazer, and Droulez}]{Coninx2016icrc}
\bibinfo{author}{A.~Coninx}, \bibinfo{author}{R.~Laurent},
  \bibinfo{author}{M.~A. Aslam}, \bibinfo{author}{P.~Bessi{\`{e}}re},
  \bibinfo{author}{J.~Lobo}, \bibinfo{author}{E.~Mazer},
  \bibinfo{author}{J.~Droulez},
\newblock \bibinfo{title}{{Bayesian Sensor Fusion with Fast and Low Power
  Stochastic Circuits}},
\newblock in: \bibinfo{booktitle}{Proceedings of the first IEEE International
  Conference on Rebooting Computing (ICRC) [In press]},
  \bibinfo{publisher}{IEEE Computer Society}, \bibinfo{address}{San Diego, CA},
  \bibinfo{year}{2016}.
\bibitem[{Elfes(1989)}]{Elfes1989}
\bibinfo{author}{A.~Elfes},
\newblock \bibinfo{title}{{Using Occupancy Grids for Mobile Robot Perception
  and Navigation}},
\newblock \bibinfo{journal}{IEEE Computer} \bibinfo{volume}{22}
  (\bibinfo{year}{1989}) \bibinfo{pages}{46--57}.
\bibitem[{Thrun(2003)}]{Thrun2003}
\bibinfo{author}{S.~Thrun},
\newblock \bibinfo{title}{{Learning occupancy grid maps with forward sensor
  models}},
\newblock \bibinfo{journal}{Autonomous Robots} \bibinfo{volume}{15}
  (\bibinfo{year}{2003}) \bibinfo{pages}{111--127}.

\end{thebibliography}

\end{document}